\PassOptionsToPackage{unicode}{hyperref}
\PassOptionsToPackage{hyphens}{url}
\documentclass[
  11pt,
]{article}
\usepackage{amsmath,amssymb}
\usepackage{iftex}
\ifPDFTeX
  \usepackage[T1]{fontenc}
  \usepackage[utf8]{inputenc}
  \usepackage{textcomp} 
\else 
  \usepackage{unicode-math} 
  \defaultfontfeatures{Scale=MatchLowercase}
  \defaultfontfeatures[\rmfamily]{Ligatures=TeX,Scale=1}
\fi
\usepackage{lmodern}
\ifPDFTeX\else
\fi
\IfFileExists{upquote.sty}{\usepackage{upquote}}{}
\IfFileExists{microtype.sty}{
  \usepackage[]{microtype}
  \UseMicrotypeSet[protrusion]{basicmath} 
}{}
\makeatletter
\@ifundefined{KOMAClassName}{
  \IfFileExists{parskip.sty}{%
    \usepackage{parskip}
  }{
    \setlength{\parindent}{0pt}
    \setlength{\parskip}{6pt plus 2pt minus 1pt}}
}{
  \KOMAoptions{parskip=half}}
\makeatother
\usepackage{xcolor}
\usepackage[margin=1in]{geometry}
\usepackage{longtable,booktabs,array}
\usepackage{calc} 
\usepackage{etoolbox}
\makeatletter
\patchcmd\longtable{\par}{\if@noskipsec\mbox{}\fi\par}{}{}
\makeatother
\IfFileExists{footnotehyper.sty}{\usepackage{footnotehyper}}{\usepackage{footnote}}
\makesavenoteenv{longtable}
\usepackage{graphicx}
\makeatletter
\def\maxwidth{\ifdim\Gin@nat@width>\linewidth\linewidth\else\Gin@nat@width\fi}
\def\maxheight{\ifdim\Gin@nat@height>\textheight\textheight\else\Gin@nat@height\fi}
\makeatother
\setkeys{Gin}{width=\maxwidth,height=\maxheight,keepaspectratio}
\makeatletter
\def\fps@figure{htbp}
\makeatother
\providecommand{\tightlist}{%
  \setlength{\itemsep}{0pt}\setlength{\parskip}{0pt}}
\usepackage{array}
\usepackage{longtable}
\usepackage{graphicx}
\usepackage{titling}
\usepackage{microtype}
\usepackage[dvipsnames]{xcolor}
\usepackage{titlesec}
\usepackage{enumitem}
\usepackage{caption}
\definecolor{brandSlate}{HTML}{0F172A}
\definecolor{brandMid}{HTML}{475569}
\definecolor{brandLink}{HTML}{0E7490}
\definecolor{brandRule}{HTML}{E2E8F0}
\usepackage{hyperref}
\hypersetup{colorlinks=true, linkcolor=brandLink, urlcolor=brandLink, citecolor=brandLink}
\titleformat{\section}{\Large\bfseries\color{brandSlate}}{\thesection}{0.6em}{}
\titleformat{\subsection}{\large\bfseries\color{brandSlate}}{\thesubsection}{0.5em}{}
\titleformat{\subsubsection}{\normalsize\bfseries\color{brandMid}}{\thesubsubsection}{0.5em}{}
\titlespacing*{\section}{0pt}{1.6ex plus 0.4ex minus 0.2ex}{0.9ex plus 0.2ex}
\titlespacing*{\subsection}{0pt}{1.2ex plus 0.3ex minus 0.2ex}{0.5ex plus 0.2ex}
\pretitle{\begin{center}\includegraphics[width=5cm]{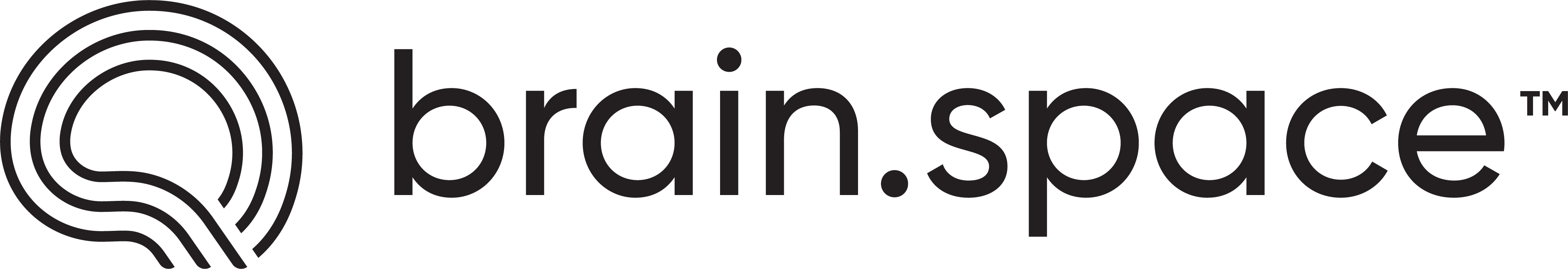}\\[1.5em]\LARGE\bfseries\color{brandSlate}}
\posttitle{\par\end{center}}
\AtBeginEnvironment{longtable}{\footnotesize}
\ifLuaTeX
  \usepackage{selnolig}  
\fi
\IfFileExists{bookmark.sty}{\usepackage{bookmark}}{\usepackage{hyperref}}
\IfFileExists{xurl.sty}{\usepackage{xurl}}{} 
\hypersetup{
  pdftitle={STST-JEPA: Shallow-Target Spatio-Temporal Joint Embedding Prediction Architecture For EEG Self-Supervised Learning},
  pdfauthor={Roy Segal (brain.space); Yoni Svechinsky (brain.space); Tomer Fekete (brain.space)},
  hidelinks,
  pdfcreator={LaTeX via pandoc}}

\title{STST-JEPA: Shallow-Target Spatio-Temporal Joint Embedding
Prediction Architecture For EEG Self-Supervised Learning}
\author{Roy Segal (brain.space) \and Yoni Svechinsky
(brain.space) \and Tomer Fekete (brain.space;
\href{mailto:tomer.fekete@brain.space}{tomer.fekete@brain.space})}
\date{}

\usepackage{newunicodechar}
\newunicodechar{−}{\ensuremath{-}}
\newunicodechar{×}{\ensuremath{\times}}
\newunicodechar{→}{\ensuremath{\rightarrow}}
\newunicodechar{↔}{\ensuremath{\leftrightarrow}}
\newunicodechar{≈}{\ensuremath{\approx}}
\newunicodechar{≥}{\ensuremath{\geq}}
\newunicodechar{≤}{\ensuremath{\leq}}
\newunicodechar{±}{\ensuremath{\pm}}
\newunicodechar{Δ}{\ensuremath{\Delta}}
\newunicodechar{λ}{\ensuremath{\lambda}}
\newunicodechar{σ}{\ensuremath{\sigma}}
\newunicodechar{κ}{\ensuremath{\kappa}}
\newunicodechar{²}{\textsuperscript{2}}
\newunicodechar{µ}{\textmu}
\newunicodechar{§}{\S}
\newunicodechar{á}{\'a}
\begin{document}
\maketitle

{
\setcounter{tocdepth}{2}
\tableofcontents
}
\hypertarget{abstract}{%
\section{Abstract}\label{abstract}}

Brain age - the age inferred from a physiological recording - is an
emerging biomarker whose deviation from chronological age tracks
neurological and psychiatric burden, and EEG is an attractive substrate
for it because it is cheap, portable, and temporally rich. Yet EEG
brain age models must contend with cross site montage heterogeneity,
small labelled cohorts, and dominant subject level nonstationarity, and
few EEG foundation models have been shown to deliver competitive age
regression across the full pediatric-to-older-adult range in which such
a biomarker would actually be deployed. We introduce \textbf{STST-JEPA},
a self-supervised transformer for resting state and task EEG, pretrained
on 47,703 sessions spanning ages 5--81 from the brain.space and Healthy
Brain Network (HBN) corpora. The model combines a latent prediction
objective - predicting masked token representations against an
EMA-of-tokenizer target - with an auxiliary signal reconstruction
term, applied to 30 second multichannel windows under spatiotemporal
block masks. A lightweight attentive probe trained on frozen pretrained
embeddings achieves a \textbf{best held out validation mean absolute
error of 3.06 years} (r = 0.924) for age regression on 3,367
sessions, against a predict-the-mean baseline of approximately 10 years
MAE. With light task specific fine tuning of the model's final layers,
the same pretrained encoder achieves rank 1 placements - with the
model's native 30 second windows - on the public NeuralBench ×
brain.space EEG leaderboard for sex classification (balanced accuracy
0.911), age prediction (r = 0.749), and psychopathology composite
regression (r = 0.215). We further show that the model's age prediction
residual is negatively correlated with cognitive efficiency over several
tasks we examined.

\hypertarget{introduction}{%
\section{Introduction}\label{introduction}}

Chronological age is a powerful anchor for resting state
electrophysiology. The EEG spectrum reorganizes systematically from
childhood through senescence - alpha peaks migrate, 1/f slopes
flatten, sleep related rhythms attenuate - and the gap between
\emph{predicted} and \emph{actual} age from these signatures has emerged
as a candidate biomarker of neurological and psychiatric burden, with
pipelines trained on healthy cohorts showing systematic residuals in
populations with cognitive decline {[}Cole \& Franke, 2017; Franke \&
Gaser, 2019{]} and in longitudinal mortality follow up from sleep EEG
{[}Sun et al., 2019{]}. EEG is a natural substrate for such a biomarker:
it is cheap, portable, and temporally rich. On the other hand, it is
resistant to the assumptions of modern deep learning architectures
(transformers in particular {[}Vaswani et al., 2017{]}): EEG montages
differ across sites (we contend with one 115 channel and one 128 channel
layout in this work alone), labeled cohorts are small relative to what
contemporary transformers expect, and subject level nonstationarity
dominates within condition variance - representations risk being tied
to the individual recording rather than to the construct the recording
is meant to express. We therefore pursue a representation that can
\emph{predict} what is missing from a recording in a space that already
abstracts over those nuisances, and that is then held accountable,
through a reconstruction term, to the waveform it came from.

Brain age prediction from M/EEG has a mature classical literature and a
fast moving deep learning one. Hand crafted spectral and covariance
features fed to ridge or random forest regressors remain strong
baselines: the benchmark of Engemann et al.~{[}2022{]}, spanning four
M/EEG cohorts and more than 2,500 participants, reports MAEs in the 7--8
year range with R² up to \textasciitilde0.75, and a best TUAB MAE of
7.75 years on cleaned adult clinical EEG. Task specific convolutional
models such as EEGNet {[}Lawhern et al., 2018{]} and the Shallow/Deep
ConvNets of Schirrmeister et al.~{[}2017{]} close much of the gap to
classical pipelines on BCI paradigms while remaining compact and
task bespoke. Large sleep EEG regressors reach MAEs near 7--8 years on
adult cohorts {[}Sun et al., 2019{]}; pediatric or narrow age studies
inevitably report smaller absolute errors that cannot be compared
directly to adult benchmarks. In parallel, EEG has acquired a first
generation of \emph{foundation models} that attempt to amortize the cost
of montage and cohort heterogeneity across an unlabeled pretraining
corpus: BENDR {[}Kostas et al., 2021{]} adapted wav2vec-style
contrastive learning to raw EEG; BrainBERT {[}Wang et al., 2023{]}
brought masked spectrogram modeling to intracranial recordings; BIOT
{[}Yang et al., 2023{]} tokenized channels independently to absorb
mismatched montages and missing channels; LaBraM {[}Jiang et al.,
2024{]} scaled to roughly 2,500 hours and hundreds of millions of
parameters with vector quantized neural spectrum prediction; Neuro-GPT
{[}Cui et al., 2024{]} trained a GPT-style masked chunk predictor on
stacked EEG; and EEG2Rep {[}Foumani et al., 2024{]} brought
joint embedding predictive training to EEG by predicting the latent
representations of masked spatiotemporal regions. Several of these
models do report age regression, and some are evaluated on shared
benchmarks such as NeuralBench; what remains comparatively rare is a
single foundation model that couples competitive \emph{age regression}
across heterogeneous, multisite corpora spanning pediatric through
older adult subjects with leaderboard level cross task transfer from one
frozen backbone - the regime in which a brain age biomarker would
actually be deployed.

Self-supervised approaches in vision span a wide design space -
latent prediction methods that match masked region representations
against an EMA target {[}iBOT, Zhou et al., 2022; LeCun, 2022; I-JEPA,
Assran et al., 2023; V-JEPA, Bardes et al., 2024; data2vec, Baevski et
al., 2022{]}, signal reconstruction approaches that predict the raw
input or its variational latent {[}VAE, Kingma \& Welling, 2013; MAE, He
et al., 2021{]}, contrastive approaches that pull augmented views of the
same input together {[}SimCLR, Chen et al., 2020; BYOL, Grill et al.,
2020{]}, and homeostatic / variance preserving regularisers {[}VICReg,
Bardes et al., 2022; LeJEPA, Balestriero \& LeCun, 2025{]}. STST-JEPA
sits in the masked prediction quadrant with an auxiliary reconstruction
term. Each family has characteristic failure modes. Pure
latent prediction can collapse absent EMA targets, stop-gradients, and
explicit regularizers - the target stream drifts toward trivial
solutions and the student tracks the drift. Pure signal reconstruction
over invests in surface statistics: on EEG, where noise, muscle
artifact, and eye blinks carry substantial energy, a decoder trained to
reproduce the waveform will, by construction, allocate capacity to
reproducing those nuisances. We instantiate a joint objective combining
the two - a pairing that maps loosely onto predictive processing
accounts of cortex {[}Rao \& Ballard, 1999; Friston, 2010; Clark,
2013{]}, in which a latent generative model is continually disciplined
by signal level consequences - but we treat that mapping as
architectural intuition rather than a claim about cortical fidelity, and
whether the joint formulation is preferable to either of its single term
ablations on this corpus is left to future work and not claimed
empirically here.

\textbf{STST-JEPA} (this paper) implements this design. The model
ingests 30 second, 256 Hz windows across a unified 128 channel budget,
collapsing per channel temporal patches at each time index with a
learned Pooled Multihead Attention (PMA) block that admits arbitrary
montages through coordinate aware, mask aware channel pooling.
Pretraining applies the latent prediction loss between predictor outputs
at masked positions and the corresponding EMA tokenizer outputs
(\(\lambda_{\mathrm{lat}} = 1.0\)), together with a smooth-L1
reconstruction loss that decodes each masked patch back to its 16 sample
waveform (\(\lambda_{\mathrm{rec}} = 0.35\)), on spatiotemporal block
masks covering \textasciitilde24\% of the grid on average. The
reconstruction term is deliberately down weighted: it is not meant to be
co-equally minimized, but to act as a soft floor that keeps the learned
latent space close enough to the signal that a short
linear--GeLU--linear head can decode it.

We examine the resulting representations on age regression over a
combined internal brain.space plus Healthy Brain Network (HBN; Alexander
et al., 2017) pretraining corpus of 47,703 sessions spanning ages 5 to
81. A lightweight attentive probe on frozen pretrained embeddings
reaches a held out validation MAE of \textbf{3.06 years} and RMSE of
\textbf{5.11 years}, against a predict-the-training-mean baseline of
10.09 years MAE / 13.27 years RMSE - a 69.7\% MAE reduction. This
result is not directly comparable to the 7--8 year MAEs reported on
cleaned adult cohorts such as TUAB: our corpus is pediatric heavy (HBN
contributes 25,115 of 47,703 sessions, with ages concentrated in
childhood and adolescence), and the conditional variance of age in such
a cohort is smaller than in adult clinical EEG, which mechanically
lowers achievable MAE. We therefore report age as a \emph{stress test}
of the representation - the cleanest yardstick for judging whether
pretraining captured subject level structure - rather than as the
paper's primary endpoint. Auxiliary probes on the same frozen embeddings
tell a complementary story: sex (balanced accuracy 0.89) and paradigm
identity (0.88 on eyes open / eyes closed, above chance on three other
task classifiers) are also recoverable. To check that this is not a
quirk of our internal split or label set, we additionally evaluate the
same pretrained checkpoint on the public \textbf{NeuralBench ×
brain.space EEG leaderboard} (final layer fine tuning per the
leaderboard protocol; Methods §\emph{External Benchmark Protocol}),
where it places rank 1 across three different downstream tasks (sex,
age, and a psychopathology composite) from a single shared backbone -
the foundation model framing of STST-JEPA is what the paper claims most
concretely, with brain age as the headline application.

We are explicit about scope. The NeuralBench × brain.space leaderboard
results (Table 4) are fixed protocol evaluations on NeuralBench's
held out test partition. For the internal brain.space cohort, the
age regression numbers we report (Table 2) are best validation values
across the training trajectory rather than a single fixed protocol test
draw; the internal test partition is reserved for that future report,
and the only place this paper touches it is the exploratory
brain age gap × behavioural capacity analysis in Results, which pools
the validation and test partitions for statistical power and is labelled
exploratory throughout. Controlled ablations of the joint objective were
beyond our compute budget, though a wide range of architectures and
objectives was explored during development. We make no clinical validity
claim. The work below should be read as an architectural and empirical
study of one design choice - joint latent-plus-signal supervision -
on one large heterogeneous corpus, reported on age regression with an
auxiliary probe panel and an external multitask benchmark.

\textbf{Contributions.}

\begin{itemize}
\tightlist
\item
  \textbf{STST-JEPA}, a 24 layer transformer foundation model with
  coordinate aware PMA channel pooling, pretrained with a joint
  latent-prediction-plus-reconstruction objective (predictor outputs
  supervised against an EMA-of-tokenizer target) on 47,703 EEG sessions
  from two cohorts with different montages and age distributions
  (brain.space and HBN), unified through a shared 128 channel budget
  with missing channel masks.
\item
  A reusable attentive probe protocol on frozen pretrained embeddings
  that supports age regression and a panel of auxiliary labels and
  paradigm probes without retraining the backbone.
\item
  Cross task generalization from a single pretrained checkpoint: rank 1
  placements on the public NeuralBench × brain.space EEG benchmark for
  sex (bal. acc. 0.911), age (Pearson r 0.749), and a psychopathology
  composite (Pearson r 0.215), reported with the model's native
  30 second windows. Under the benchmark's standard 2 second protocol,
  sex and psychopathology remain rank 1 while age is rank 4 (§External
  Benchmark Performance). This is direct evidence of the model's
  \emph{foundation model} status - one backbone, three diverse
  downstream tasks.
\item
  Brain age as the headline application: best held out validation
  \textbf{3.06 yr MAE} and \textbf{5.11 yr RMSE} on a subject disjoint
  partition spanning 5--81 years, complemented by an auxiliary probe
  panel showing that the same frozen embeddings also recover sex and
  paradigm identity well above chance.
\item
  An exploratory BAG × behavioural capacity analysis on the union of the
  held out validation and test partitions: 7 of 21 cognitive efficiency
  targets survive a Benjamini--Hochberg FDR correction (q = 0.05), all
  in the expected negative direction (older looking brain ↔ worse
  efficiency); effect sizes are small (\textbar r\textbar{} \textless{}
  0.10), and the load bearing observation is direction-of-effect
  consistency rather than effect size discovery.
\item
  A brain inspired framing that motivates the joint objective by loose
  analogy to predictive processing accounts of cortex (architectural
  intuition, not a claim about cortical fidelity).
\end{itemize}

\hypertarget{methods}{%
\section{Methods}\label{methods}}

We pretrain an EEG foundation model - hereafter \textbf{STST-JEPA} -
using a latent prediction objective (predicting masked token
representations against an EMA-of-tokenizer target) combined with a
per patch signal reconstruction term, and we evaluate the learned
representations on downstream age regression. The remainder of this
section describes the pretraining corpus and cohort splits, the EEG
preprocessing pipeline, the tokenizer and encoder architecture, the
self-supervised objective, optimization, and the attentive age probe.
All values below refer to the configuration used to train the reported
run; a compact hyperparameter summary is provided in Table 1.

\hypertarget{datasets-and-cohort-partitioning}{%
\subsection{Datasets and Cohort
Partitioning}\label{datasets-and-cohort-partitioning}}

Pretraining uses a combined EEG corpus drawn from two sources: an
internal brain.space corpus and the publicly available Healthy Brain
Network (HBN) dataset. The corpus contains 47,703 sessions in total,
partitioned at the subject level into 67.6\% train (32,246 sessions),
12.2\% validation (5,802 sessions), and 20.2\% held out test (9,655
sessions); of these, 22,588 originate from brain.space and 25,115 from
HBN. The test partition is deliberately larger than validation because
an entire task paradigm was held out within it for later
experimentation. The age distribution of the training partition spans 5
to 81 years (mean 17.7 ± 12.6 years; median \textasciitilde14 years),
dominated by pediatric and young adult recordings from HBN and extending
through middle and older adulthood via brain.space.

To prevent information leakage across repeated recordings, partitioning
is performed at the subject level and then propagated to session
identifiers. Splits are generated independently within each data source
and stratified by a composite label formed from (i) the number of
available sessions per subject (capped at 5), (ii) a discretized age bin
(10 bins for brain.space, 12 for HBN), and (iii) a source specific
demographic covariate - handedness for brain.space and sex for HBN.
Subjects are assigned to train, validation, and test partitions using
\texttt{sklearn.train\_test\_split} with \texttt{random\_state=42}, and
the resulting subject level split is propagated to each of their
sessions. This procedure preserves source balance and age coverage
across partitions while eliminating the risk that multiple recordings
from the same individual appear in different partitions.

\textbf{Table 0 - Pretraining cohort summary (session counts by split
× source, plus age statistics).} Ages are pooled across sources within
each split.

\begin{longtable}[]{@{}
  >{\raggedright\arraybackslash}p{(\columnwidth - 10\tabcolsep) * \real{0.1364}}
  >{\raggedleft\arraybackslash}p{(\columnwidth - 10\tabcolsep) * \real{0.1818}}
  >{\raggedleft\arraybackslash}p{(\columnwidth - 10\tabcolsep) * \real{0.1818}}
  >{\raggedleft\arraybackslash}p{(\columnwidth - 10\tabcolsep) * \real{0.1818}}
  >{\raggedright\arraybackslash}p{(\columnwidth - 10\tabcolsep) * \real{0.1364}}
  >{\raggedleft\arraybackslash}p{(\columnwidth - 10\tabcolsep) * \real{0.1818}}@{}}
\toprule\noalign{}
\begin{minipage}[b]{\linewidth}\raggedright
Split
\end{minipage} & \begin{minipage}[b]{\linewidth}\raggedleft
Sessions (total)
\end{minipage} & \begin{minipage}[b]{\linewidth}\raggedleft
brain.space
\end{minipage} & \begin{minipage}[b]{\linewidth}\raggedleft
HBN
\end{minipage} & \begin{minipage}[b]{\linewidth}\raggedright
Age mean ± sd (yr)
\end{minipage} & \begin{minipage}[b]{\linewidth}\raggedleft
Age range (yr)
\end{minipage} \\
\midrule\noalign{}
\endhead
\bottomrule\noalign{}
\endlastfoot
Train & 32,246 & 15,277 & 16,969 & 17.7 ± 12.6 & 5--81 \\
Validation & 5,802 & 2,777 & 3,025 & 17.6 ± 12.6 & 5--81 \\
Test & 9,655 & 4,534 & 5,121 & 17.7 ± 12.6 & 5--81 \\
\textbf{Total} & \textbf{47,703} & \textbf{22,588} & \textbf{25,115} &
\textbf{17.7 ± 12.6} & \textbf{5--81} \\
\end{longtable}

The brain.space portion of the pretraining cohort additionally exposes a
curated brain age finetune cohort (9,599 brain.space sessions / 2,548
subjects, ages 18--81, mean 30.8 ± 13.9 yr) used as the downstream
evaluation cohort; a next version cohort of roughly 20,000 sessions is
currently in training. The brain age gap × behaviour analysis in
Brain Age Gap and Behavioural Performance uses the union of this
cohort's validation and test partitions (N = 8,600 sessions, 2,109
subjects). HBN does not participate in any downstream evaluation
reported here.

\hypertarget{eeg-segmentation-and-preprocessing}{%
\subsection{EEG Segmentation and
Preprocessing}\label{eeg-segmentation-and-preprocessing}}

Pretraining examples are drawn from two recording paradigms common to
both corpora: a multi-object-tracking task (\texttt{mot}) {[}Pylyshyn \&
Storm, 1988{]} and an eyes open / eyes closed resting block
(\texttt{eo\_ec}). Each example is a \(W_{\mathrm{win}} = 30.0\)s window
sampled at \(f_s = 256\) Hz, giving
\(S = W_{\mathrm{win}} \cdot f_s = 7,680\) samples per channel. Windows
are extracted with a 6.0s stride during training and a 2.0s stride
during validation to increase the effective density of validation
examples without inflating training redundancy.

The two corpora use different electrode montages - brain.space
recordings have 115 channels and HBN recordings have up to 128 channels
- so the model operates on a unified channel budget of \(C = 128\)
with per channel presence indicators. Missing channels are padded with
zeros and flagged in a sample level validity mask; the padded slots
participate in all downstream tensor operations but are suppressed
inside the input embedder via the same mask.

For each window the dataloader constructs a binary validity mask from
three sources: (i) missing channel indicators carried in the metadata,
(ii) invalid channel statistics (missing or nonfinite channel median or
interquartile range), and (iii) an amplitude based artifact rule that
masks any sample whose absolute value exceeds 150 (in the recording's
native amplitude units) and expands the mask by a one sample temporal
buffer to suppress partial transients. Windows in which the combined
invalid fraction exceeds 45\% are rejected and resampled (up to five
retries). Valid channels are then robustly normalized using channel wise
medians and interquartile ranges,

\[
\tilde{x}_{c,s} = \frac{x_{c,s} - \mathrm{median}_c}{\mathrm{IQR}_c},
\]

where \(c\) indexes channels and \(s\) indexes time samples; invalid
channels are set to zero in the numerator and unit scale in the
denominator, rendering them identically zero after normalization. Each
recording is accompanied by a source specific 3D electrode coordinate
template (a distinct template for brain.space and HBN) so that the two
montages are expressed in a common spatial frame while preserving their
respective geometries.

\textbf{Upstream signal conditioning.} The SSL pipeline consumes
already conditioned ICA cleaned data and does \emph{not} perform
additional band pass, line noise, or re-referencing operations at
training time. Upstream of the SSL pipeline, both corpora receive
analogous conditioning at source acquisition rate - notch filtering,
artifact subspace reconstruction (ASR), and ICA. Concretely, the
continuous EEG is high pass filtered at 1 Hz (first order) with
transient spikes interpolated out, then notch filtered at the line
frequency and its harmonics (order 10 band stops) at the 500 Hz
acquisition rate, followed by three artifact stages - bad channel
removal, ASR, and ICA. \textbf{Bad channel removal} is the main point of
divergence: brain.space uses a per period spectral signal quality model
with a Mahalanobis outlier test (a channel is rejected if it is
spectrally degenerate or its five dimensional spectral feature vector
- dropped window fraction, high frequency tail power, aperiodic
\(1/f\) slope, residual line power, and variance about the \(1/f\) fit
- lies beyond a fixed reference distribution; recordings with more
than 20 bad channels fail QC), whereas HBN, lacking event structure,
applies a fixed absolute RMS threshold (RMS \(>\) 60 µV) and always
drops the reference channel. Bad channels are excluded from the ASR/ICA
computation in both corpora; brain.space masks and reinserts them so the
full montage is preserved, while HBN drops their columns. \textbf{ASR}
learns a robust (shrinkage) covariance of the good channels on a
calibration segment and, in 0.5 s sliding windows, reconstructs any
principal component exceeding \(\kappa = 20\) robust standard deviations
from the retained clean subspace - calibrated on the eyes closed rest
period for brain.space and on the whole session for HBN. \textbf{ICA}
re-references the ASR cleaned good channels to the common average and
decomposes them with the Picard algorithm (extended, nonorthogonal)
into up to 42 components (capped at the number of surviving good
channels); ocular components (spatially frontal, with mean absolute
frontal mixing weight above a robust upper tail threshold) and cardiac
components (a 0.6--1.7 Hz peak confirmed by a cardiac influence factor
and a beat shape correlation) are projected out, while components
exhibiting a resting alpha peak (8--12 Hz power exceeding 5--8 Hz) are
protected as neural. Only the confirmed ocular and cardiac components
are removed; the SSL pipeline itself adds no further band pass,
line noise, or re-referencing (the HBN input corresponds to the
published \texttt{hbn-eeg-post-asr-ica} derivative). The reference
scheme used by the model is whatever was applied by the source pipeline
(no global re-referencing layer is inserted by the SSL pipeline); the
model is exposed to residual between cohort heterogeneity through the
per source coordinate template and the unified channel budget rather
than through an explicit harmonization step. After ICA, all recordings
are resampled (where needed) to a common 256 Hz via polyphase resampling
(\texttt{scipy.signal.resample\_poly}) prior to window extraction. The
``native amplitude units'' referenced in the amplitude rejection rule
are the source pipeline's post ICA microvolt scale floats; the model
itself never sees raw microvolts because every channel is
robust-z-scored per window before tokenization, so the optimizer's
effective input scale is unit-IQR rather than µV.

\hypertarget{tokenization-and-input-embedding}{%
\subsection{Tokenization and Input
Embedding}\label{tokenization-and-input-embedding}}

Let an input window after preprocessing be
\(X \in \mathbb{R}^{C \times S}\) with \(C = 128\) and \(S = 7,680\).
Each channel is patchified independently by a weight shared 1-D temporal
convolution with patch length \(P = 16\) and stride \(\Delta = 16\),
yielding

\[
T = \left\lfloor \frac{S - P}{\Delta} \right\rfloor + 1 = 480
\]

temporal patches per channel and \(C \cdot T = 61,440\) candidate
spatiotemporal patches per example. Expanding the token grid to this
size directly would make attention over all channel time locations
prohibitively expensive, so we collapse the channel axis at every
temporal index with a learned PMA module.

Specifically, at each temporal index \(t\) the \(C\) per channel patch
embeddings are integrated into a single \(d\)-dimensional token by a
Pooled Multihead Attention (PMA) block {[}Set Transformer, Lee et al.,
2019{]} in which \(Q = 16\) learned \emph{inducing queries} attend over
the channel dimension key/value set via \(H = 16\)-head cross-attention.
Prior to pooling, each per channel patch embedding is combined with a
learned projection of that channel's 3D electrode coordinate, modulated
by a learnable scalar gate \(g_{\mathrm{coord}}\) initialized to zero
and trained at a 5× learning rate; this allows the model to begin
training in a coordinate free regime and progressively incorporate
montage geometry as representations stabilize. Invalid or missing
channels are suppressed inside the PMA pooling via the sample level
validity mask, so that tokens are never contaminated by padded or
artifact rejected channels. The \(Q\) pooled queries are concatenated
and linearly projected to the model width \(d = 768\), producing one
token per temporal index. Temporal position is injected inside the
transformer attention layers via rotary position embeddings {[}RoPE, Su
et al., 2021{]} at base \(10,000\); no additive time position embedding
is used. The resulting token sequence has length \(T = 480\) and width
\(d = 768\).

\hypertarget{encoder-target-tokenizer-and-predictor}{%
\subsection{Encoder, Target Tokenizer, and
Predictor}\label{encoder-target-tokenizer-and-predictor}}

The architecture has three components: a \textbf{lightweight tokenizer},
a \textbf{deep context encoder}, and a \textbf{predictor}. Targets for
the self-supervised loss are produced by an exponential moving average
of the \textbf{tokenizer} (not of the deep encoder); the deep encoder is
one sided.

\textbf{Tokenizer.} The per channel temporal convolution described in
\emph{Tokenization and Input Embedding} (patch length 16, stride 16,
plus the PMA channel pool and coordinate / time position embeddings) is
the tokenizer. It produces the \(T = 480\)-token sequence that all
downstream modules consume.

\textbf{Context encoder.} A prenormalized {[}Xiong et al., 2020{]}
transformer stack of 24 layers, 16 heads, FFN inner dimension 3,072,
GeLU, dropout 0.1, operating at width 768. It receives the visible
(unmasked) tokens from the tokenizer and produces deep contextual
representations for the predictor.

\textbf{Predictor.} A two layer transformer with cross-attention that
consumes the context encoder's visible token representations as
keys/values, sinusoidal position queries at the masked positions, and
produces a predicted representation at each masked position. RoPE is
disabled inside the predictor so that target position is supplied only
through the sinusoidal query tokens. The predictor output is read off at
the last layer before its final cross-attention step (the ``predicted
token'') and passed through a nonlinear projection head into the loss
space, \[
h_{\mathrm{proj}}(z) = W_2\,\sigma\!\bigl(W_1\,\mathrm{LN}(z)\bigr),
\] with \(W_1, W_2 \in \mathbb{R}^{d \times d}\), \(d = 768\), and
\(\sigma = \mathrm{GeLU}\). The projection is applied \textbf{only on
the predictor side}; EMA tokenizer targets enter the loss in their
native embedding space (also \(\mathbb{R}^d\)) without a projection
head. Its role is to give the predictor a small adapter into the
target token space rather than to change dimensionality.

\textbf{EMA target stream.} A parameter frozen copy of the tokenizer is
maintained as an exponential moving average of the live tokenizer. Its
parameters are excluded from gradient flow (stop-gradient), as in BYOL
{[}Grill et al., 2020{]} and JEPA-family training {[}Assran et al.,
2023; Bardes et al., 2024{]}, and dropout is disabled. The EMA momentum
\(m\) is linearly warmed from \(m_{\mathrm{min}} = 0.9996\) to
\(m_{\mathrm{max}} = 0.9999\) over the first 12,000 optimizer steps and
held at \(m_{\mathrm{max}}\) thereafter:

\[
\theta_{\mathrm{tgt}} \leftarrow m\,\theta_{\mathrm{tgt}} + (1-m)\,\theta_{\mathrm{tok}}.
\]

The full input window (unmasked) is tokenized by the EMA tokenizer to
produce target tokens at every (channel, time) position; the
self-supervised loss compares predictor outputs at masked positions to
these EMA tokenizer targets at the same positions. There is no EMA twin
of the deep context encoder - the target stream is
\textbf{tokenizer deep, not encoder deep}.

\textbf{Relation to JEPA and MAE.} This design borrows JEPA-family
training discipline - an EMA target with stop-gradient and a separate
predictor - but applies it at the \textbf{tokenizer level} rather than
at the full encoder level. Compared to canonical JEPA, the predicted
representations are shallow (tokenizer outputs), not the deep encoder's
late layer features. Compared to canonical MAE, the loss does not
require a heavy decoder back to raw signal and instead reconstructs the
online (EMA) tokenizer's own outputs (a lightweight raw signal
reconstruction term is layered on separately; see below). The
construction can equivalently be read as a \textbf{shallow target JEPA}
or an \textbf{intermediate latent MAE with EMA targets}; the closest
published analogue we are aware of is a multilayered MAE with
EMA targeted tokens.

\hypertarget{self-supervised-objective}{%
\subsection{Self-Supervised Objective}\label{self-supervised-objective}}

Pretraining combines a masked token prediction loss in tokenizer output
space with a lightweight per patch signal reconstruction loss.

\textbf{Masking.} For each example, a spatiotemporal mask

\[
\mathcal{M} \subseteq \{1,\dots,C\} \times \{1,\dots,T\}
\]

is sampled as the union of 4 rectangular blocks on the channel time
grid. Each block independently covers a random fraction of channels
drawn uniformly from \([0.2,\,0.5]\) and a random fraction of temporal
patches drawn uniformly from \([0.1,\,0.3]\), with the block's channel
and time footprints sampled independently. Because blocks may overlap,
the realized mask ratio is stochastic; empirically it concentrates near
24\% of grid positions during training.

\textbf{Latent prediction loss.} Visible tokens are processed by the
context encoder and then by the predictor, which emits a predicted
representation \(\hat{z}_i\) at each masked position
\(i \in \mathcal{M}\) (passed through the predictor side projection head
defined above). In parallel, the EMA tokenizer ingests the \emph{full}
(unmasked) input window and emits a target representation
\(z_i^{\mathrm{tgt}}\) at every (channel, time) position. The
latent prediction loss is a mean squared error in the loss space,

\[
\mathcal{L}_{\mathrm{lat}}
=
\frac{1}{|\mathcal{M}|}
\sum_{i \in \mathcal{M}}
\bigl\|\,\operatorname{sg}(z_i^{\mathrm{tgt}}) - \hat{z}_i\,\bigr\|_2^2,
\]

where \(\operatorname{sg}(\cdot)\) denotes stop-gradient through the
EMA tokenizer branch.

\textbf{Per patch reconstruction loss.} A lightweight two layer
reconstruction head
\(g_{\mathrm{rec}}: \mathbb{R}^{768} \to \mathbb{R}^{P}\) with
architecture LayerNorm → Linear(\(768 \to 1{,}536\)) → GeLU →
Linear(\(1{,}536 \to 16\)) maps each masked patch prediction back to its
raw \(P=16\)-sample temporal patch. The reconstruction loss is a Huber
(smooth-L1) term with \(\beta = 1.0\),

\[
\mathcal{L}_{\mathrm{rec}}
=
\frac{1}{|\mathcal{M}|}
\sum_{i \in \mathcal{M}}
\mathrm{SmoothL1}_{\beta=1}\!\bigl(g_{\mathrm{rec}}(\hat{z}_i),\,X^{\mathrm{patch}}_i\bigr),
\]

evaluated only at those masked positions that are also marked valid by
the preprocessing mask (i.e.~excluding artifact suppressed or padded
channels). Mask coordinates \(\mathcal{M}\) and reconstruction targets
\(X^{\mathrm{patch}}_i\) are both indexed on the prepooling (channel,
time) patch grid.

\textbf{Combined objective.} The full pretraining loss is the weighted
sum

\[
\mathcal{L}
=
\lambda_{\mathrm{lat}}\,\mathcal{L}_{\mathrm{lat}}
+
\lambda_{\mathrm{rec}}\,\mathcal{L}_{\mathrm{rec}},
\]

with \(\lambda_{\mathrm{lat}} = 1.0\) and
\(\lambda_{\mathrm{rec}} = 0.35\).

\hypertarget{optimization}{%
\subsection{Optimization}\label{optimization}}

Parameters are optimized with AdamW using a peak learning rate of
\(5 \times 10^{-5}\), weight decay 0.05, and global norm gradient
clipping at 1.0. Layer normalization parameters are exempted from weight
decay; the coordinate embedding gate and related gating parameters are
trained with a 5× larger learning rate and zero weight decay, allowing
them to move quickly without contributing to the regularization budget.
The learning rate follows a cosine-warmup-to-constant schedule: starting
at one tenth of the peak LR (\(5 \times 10^{-6}\)), warming linearly
over 3,000 optimizer steps to \(5 \times 10^{-5}\), then held constant
for the remainder of training.

Training is implemented in PyTorch with PyTorch Lightning and runs in
\texttt{16-mixed} automatic mixed precision, distributed across 7 GPU
workers via PyTorch DDP. The per device batch size is 35 windows,
yielding a global batch of 245 windows (\(\approx 117,600\) patch tokens
per optimizer step). Models are compiled with \texttt{torch.compile}.
Seed 42 is fixed for data sampling, mask generation, and subject
partitioning. The reported results correspond to training through 15
epochs (≈ 155,500 optimizer steps).

\hypertarget{downstream-age-prediction}{%
\subsection{Downstream Age Prediction}\label{downstream-age-prediction}}

Age is evaluated with a lightweight attentive regression probe trained
on precomputed pretrained embeddings. During self-supervised training,
per window token sequences at the model embedding dimension
\(d_p = 768\) are calculated per epoch. The token sequences are used for
monitoring the training and for model evaluation on downstream tasks.

The probe operates at \(d_p\) and stacks three blocks: (i) a
\textbf{self-attention block} that ingests the per window token sequence
with an additional learned \texttt{CLS} token prepended (prenormalized
multihead self-attention with residual); (ii) a \textbf{single query
cross-attention block} in which the \texttt{CLS} token attends to the
rest of the sequence (prenormalized multihead cross-attention with
residual + position wise MLP); and (iii) a \textbf{linear regression
head} (LayerNorm → Dropout → Linear → 1) that decodes a scalar age from
the updated \texttt{CLS} token. Dropout is 0.1 throughout. The
single query cross-attention design reduces the post attention MLP from
\(O(T\cdot d_p)\) parameters to \(O(d_p)\), and the architecture imposes
no assumption on sequence length, so the same probe can be reused if
window length or token count are varied.

To keep probe training balanced across sources and subjects, we apply
two caps when assembling the probe's training set: for each age label
the per session window count is capped at 5, and for HBN specifically at
most 5 windows are retained per session (both caps seeded
deterministically). Train side sessions are further split internally
with a 0.2 validation fraction (seeded) for probe level early stopping.
The probe is trained with AdamW at learning rate \(10^{-3}\) for up to
15 epochs with \textbf{early stopping patience 3} (training halts when
the internal validation MAE fails to improve for 3 consecutive epochs).
The probe is optimised at the \emph{window} level - each window
receives its own smooth-L1 (Huber) loss against the session's age label
during training; window predictions are averaged at evaluation time (see
equation below).

Let \(\hat{y}_{m,j}\) denote the probe's age prediction on the \(j\)-th
window of session \(m\). Window level predictions are averaged within
each session before scoring,

\[
\hat{y}_m = \frac{1}{n_m}\sum_{j=1}^{n_m} \hat{y}_{m,j},
\]

and performance is reported as mean absolute error and root mean squared
error over session level predictions on the held out validation
partition,

\[
\mathrm{MAE} = \frac{1}{M}\sum_{m=1}^{M} \bigl| \hat{y}_m - y_m \bigr|,
\qquad
\mathrm{RMSE} = \sqrt{\frac{1}{M}\sum_{m=1}^{M} \bigl( \hat{y}_m - y_m \bigr)^2},
\]

where \(M\) is the number of evaluated sessions.

\hypertarget{auxiliary-probe-protocol}{%
\subsection{Auxiliary Probe Protocol}\label{auxiliary-probe-protocol}}

The auxiliary probes summarized in Results (Table 3) share the same
self-attention → single query cross-attention → linear head architecture
as the age probe described above, with task specific output heads: the
linear regression head is replaced by a linear classifier emitting class
logits, and the loss is multiclass cross-entropy with
\texttt{compute\_class\_weight(\textquotesingle{}balanced\textquotesingle{})}-style
per class weights to neutralize majority class skew during training
(matching the balanced accuracy reporting metric). Probe inputs are the
pretrained embeddings produced at the same per epoch evaluation cadence
used for age. Class label cardinalities are inferred from the dataset
manifest (sex 2; eyes open/closed 2; n-back 3; multi-object-tracking 6;
multimodal sensory task 6). The same per session window cap as the age
probe (5, raised to 15 partway through training) applies. Classifiers
are scored with \textbf{balanced accuracy}
(\texttt{sklearn.metrics.balanced\_accuracy\_score}), insensitive to
class imbalance.

\hypertarget{external-benchmark-protocol}{%
\subsection{External Benchmark
Protocol}\label{external-benchmark-protocol}}

The NeuralBench × brain.space EEG leaderboard evaluation (Table 4)
follows the leaderboard's published protocol. For each task we
initialise from the same pretrained model and \textbf{fine tune the
model's final layers} (encoder lower layers held fixed) on the
leaderboard's training partition; the leaderboard's held out test
partition is the evaluation set. Reported metrics are
\texttt{test/bal\_acc} for sex and \texttt{test/pearsonr} for age and
psychopathology. The leaderboard's training partition is the same
partition we drew our pretraining data from on the HBN derived portion
of NeuralBench; because the leaderboard's test partition is held out by
NeuralBench's benchmark construction (its train and test partitions are
subject disjoint by design) and was never touched by our pretraining,
this is the equivalent of using the leaderboard's intended train/test
split.

\hypertarget{brain-age-gap-bag-construction}{%
\subsection{Brain Age Gap (BAG)
Construction}\label{brain-age-gap-bag-construction}}

For the behavioural validity analysis (Results: Brain Age Gap and
Behavioural Performance) we use the \emph{bias corrected} brain age gap,
computed on the union of the brain.space finetune cohort's validation
and test partitions (N = 8,600 sessions, 2,109 subjects). The raw
prediction \(\hat{y}\) on this union is regressed on chronological age
\(y\) in a 5 fold cross-fit with \texttt{random\_state=42}: in each fold
the OLS slope \(\alpha\) and intercept \(\beta\) are estimated on the
training folds, and the corrected residual is \[
\mathrm{BAG}_i = \hat{y}_i - (\alpha\,y_i + \beta).
\] Reported coefficients are the mean ± sd across folds
(\(\alpha = 0.642 \pm 0.006\), \(\beta = 8.65 \pm 0.15\)). The cross-fit
is necessary to keep the bias correction model honest with respect to
the same union it predicts on. A quadratic sensitivity variant fits a
second order polynomial in \(y\) on the same folds; the resulting
predictions differ from the linear correction by at most
\(|\Delta r| = 0.010\) across the full 21 target panel (Results), so the
linear correction is treated as primary and the quadratic as a
robustness check. By construction the corrected \(\mathrm{BAG}\) is
orthogonal to chronological age in the OLS sense (sample
\(\mathrm{corr}(\mathrm{BAG}, y) \approx 4 \times 10^{-5}\) on the
union), so any second stage partialling on age is a no-op.

\hypertarget{behavioural-capacity-targets-and-covariates}{%
\subsection{Behavioural Capacity Targets and
Covariates}\label{behavioural-capacity-targets-and-covariates}}

The 21 behavioural capacity targets are drawn from a brain.space
production task battery (n-back, Stroop, Flanker,
multiple-object-tracking {[}MOT{]}, P3 oddball, dual oddball, and a
multimodal sensory task {[}MMST{]}). Each task family contributes a
per session \textbf{efficiency score}, a weighted combination of
accuracy and reaction-time-correct that captures speeded cognitive
performance for that family on that recording. Per family efficiency
scores are then aggregated per subject (mean over completed task
families with a minimum-of-two-families rule) and robust-z-scored to
give \emph{capacity composites}.

The 21 targets arrive at two grains:

\begin{itemize}
\tightlist
\item
  \textbf{Subject level} (14 targets): per subject capacity composites
  - Stroop efficiency, n-back efficiency, Flanker efficiency, MOT
  efficiency, MMST efficiency, dual oddball efficiency, P3 efficiency;
  plus higher order composites: \textbf{general efficiency} (mean of
  available family scores), \textbf{age residualised general efficiency}
  (general efficiency minus its age regression fit; the cleanest
  ``beyond age'' target), \textbf{attention control} (mean of Flanker,
  Stroop, P3, dual oddball), \textbf{working memory} (mean of n-back and
  n-back load resilience), \textbf{load resilience} (high load vs
  low load contrast across n-back and MOT), \textbf{stress resilience}
  (MMST i3 vs i0 contrast), and \textbf{speed--accuracy policy} (a
  fast-vs-accurate phenotype axis). Each value is broadcast to all of
  that subject's recordings at correlation time.
\item
  \textbf{Per recording} (7 targets): the per task efficiency scores
  themselves (Stroop, n-back, Flanker, MOT, MMST, dual, P3) at their
  native one-value-per-recording granularity, no z-scoring or
  aggregation.
\end{itemize}

For each (BAG, target) pair we report Pearson \(r\) as the headline
statistic, the corresponding two sided Fisher-\(z\) statistic, and a
\emph{partial} Pearson \(r\) controlling for a signal quality nuisance
bundle: \texttt{time\_of\_day\_hours}, head \texttt{circumference},
electrode distance offsets (\texttt{dist\_nasin},
\texttt{dist\_temtem}), \texttt{hair\_type}, \texttt{hair\_wash},
\texttt{glasses} use, and \texttt{handedness}. Per the bundle's design
(Memory: ``age is causal substrate, not a confound'') we do \textbf{not}
include age, sex, sleep, or education in the partial r covariate bundle:
these are causal substrates of cognitive performance rather than
nuisance confounds, and partialling them out would suppress signal that
legitimately propagates through the brain age channel. Subject level
n-inflation (one \(z\)-score broadcast across that subject's sessions)
is reported transparently and the per session rows are flagged as the
honest-n floor.

\hypertarget{multiple-comparison-handling}{%
\subsection{Multiple Comparison
Handling}\label{multiple-comparison-handling}}

The headline statistic on the 21 target panel is Benjamini--Hochberg FDR
at \(q = 0.05\) on the raw \(p\)-values (two sided) derived from the
per target Fisher-\(z\). Seven of the 21 targets survive BH-FDR at this
threshold: subject level Stroop, n-back, general efficiency,
age residualised general efficiency, attention control, and P3
efficiency, plus the per recording Stroop. We make no claim of
independence across targets - subject level and per recording Stroop
share trial level data, and several capacity composites share
constituent task families - so the BH pass count should be read as a
coarse calibration rather than a fully independent multiple comparison
test.

\hypertarget{hyperparameter-summary}{%
\subsection{Hyperparameter Summary}\label{hyperparameter-summary}}

\textbf{Table 1 - Key hyperparameters.}

\begin{longtable}[]{@{}
  >{\raggedright\arraybackslash}p{(\columnwidth - 4\tabcolsep) * \real{0.3333}}
  >{\raggedright\arraybackslash}p{(\columnwidth - 4\tabcolsep) * \real{0.3333}}
  >{\raggedright\arraybackslash}p{(\columnwidth - 4\tabcolsep) * \real{0.3333}}@{}}
\toprule\noalign{}
\begin{minipage}[b]{\linewidth}\raggedright
Group
\end{minipage} & \begin{minipage}[b]{\linewidth}\raggedright
Setting
\end{minipage} & \begin{minipage}[b]{\linewidth}\raggedright
Value
\end{minipage} \\
\midrule\noalign{}
\endhead
\bottomrule\noalign{}
\endlastfoot
Input & Window length \(W_{\mathrm{win}}\) & 30.0s \\
& Sample rate \(f_s\) & 256 Hz \\
& Samples per channel \(S\) & 7,680 \\
& Channels \(C\) & 128 \\
& Train stride & 6.0s \\
& Val stride & 2.0s \\
& Artifact amplitude threshold & 150 (native units) \\
& Max invalid fraction per window & 0.45 \\
Tokenizer & Patch length \(P\) & 16 samples \\
& Temporal patches per channel \(T\) & 480 \\
& Inducing queries \(Q\) & 16 \\
& PMA heads & 16 \\
Encoder & Embedding width \(d\) & 768 \\
& Encoder layers & 24 \\
& Attention heads & 16 \\
& FFN inner dim & 3,072 \\
& Dropout & 0.1 \\
& Positional encoding & RoPE (base 10,000) \\
Predictor & Layers & 2 \\
Projection (predictor side only) & Output width & 768 \\
& Hidden width & 768 \\
EMA & \(m_{\mathrm{min}} \to m_{\mathrm{max}}\) &
\(0.9996 \to 0.9999\) \\
& Momentum warmup & 12,000 steps \\
Masking & Blocks per example & 4 \\
& Per block channel fraction & \(\mathcal{U}[0.2, 0.5]\) \\
& Per block time fraction & \(\mathcal{U}[0.1, 0.3]\) \\
Loss & \(\lambda_{\mathrm{lat}}\) (latent prediction, MSE) & 1.0 \\
& \(\lambda_{\mathrm{rec}}\) (smooth-L1, \(\beta = 1\)) & 0.35 \\
Optimization & Optimizer & AdamW \\
& Peak LR & \(5 \times 10^{-5}\) \\
& LR schedule & cosine-warmup-to-constant, 3,000 step warmup, start
factor 10 \\
& Weight decay & 0.05 \\
& Gate LR multiplier & 5× \\
& Gradient clip (global norm) & 1.0 \\
& Precision & 16-mixed \\
& Devices & 7 (DDP) \\
& Batch per device / global & 35 / 245 \\
& \texttt{torch.compile} & enabled \\
& Seed & 42 \\
Probe & Architecture & self-attention (CLS) + single query
cross-attention + linear head \\
& Probe width \(d_p\) & 768 \\
& Max epochs / patience & 15 / 3 \\
& Learning rate & \(10^{-3}\) \\
& Internal val fraction & 0.2 \\
& Per label / HBN windows-per-session cap & 5 / 5 \\
& Eval cadence & per epoch \\
\end{longtable}

\hypertarget{results}{%
\section{Results}\label{results}}

Tables 2 and 3 report results on the held out validation partition of
the subject level split described in Methods (N = 3,367 sessions). Table
4 uses the NeuralBench external test protocol. Table 5 (Brain Age Gap
behavioural validation, an exploratory analysis) is computed on the
union of validation and test partitions (N = 8,600 sessions, 2,109
subjects) - this is the only place in the paper that draws on our
internal held out test partition, and the analysis there should be read
as exploratory rather than as a fixed protocol test set evaluation (the
NeuralBench results in Table 4 are, separately, fixed protocol
evaluations on NeuralBench's own held out test partition). Downstream
probes were trained from scratch at each evaluation rather than
fine tuned.

Combined pretraining wall clock across the two runs was approximately 79
h on 7 GPU workers (DDP, \texttt{16-mixed} precision). All downstream
probe evaluations were performed at the per epoch cadence defined in
Methods.

\hypertarget{validation-age-regression}{%
\subsection{Validation Age Regression}\label{validation-age-regression}}

The attentive probe reaches a \textbf{best held out validation} MAE of
\textbf{3.06 years} and RMSE of \textbf{5.11 years} on the 3,367
held out validation sessions. Table 2 summarizes this against a
predict-the-training-mean baseline. The 3.06 year figure is the
\emph{best validation MAE across the model's training trajectory}; it is
not a fixed protocol test set value. It is also the \textbf{combined
brain.space + HBN} validation MAE (Figure 1b): HBN is pediatric heavy
and pediatric age is comparatively easy to decode, so the combined value
(3.06 yr) is lower than the brain.space only validation MAE of 4.82 yr
(RMSE 7.06 yr, R² 0.654; Table 2). For transparency we report both
grains, noting that the NeuralBench age evaluation (Table 4) is reported
on HBN.

\textbf{Table 2 - Age prediction on the held out validation set
(combined brain.space + HBN, N = 3,367 sessions).}

\begin{longtable}[]{@{}
  >{\raggedright\arraybackslash}p{(\columnwidth - 6\tabcolsep) * \real{0.6216}}
  >{\raggedleft\arraybackslash}p{(\columnwidth - 6\tabcolsep) * \real{0.1351}}
  >{\raggedleft\arraybackslash}p{(\columnwidth - 6\tabcolsep) * \real{0.1486}}
  >{\raggedleft\arraybackslash}p{(\columnwidth - 6\tabcolsep) * \real{0.0946}}@{}}
\toprule\noalign{}
\begin{minipage}[b]{\linewidth}\raggedright
Method
\end{minipage} & \begin{minipage}[b]{\linewidth}\raggedleft
MAE (yr)
\end{minipage} & \begin{minipage}[b]{\linewidth}\raggedleft
RMSE (yr)
\end{minipage} & \begin{minipage}[b]{\linewidth}\raggedleft
R²
\end{minipage} \\
\midrule\noalign{}
\endhead
\bottomrule\noalign{}
\endlastfoot
STST-JEPA + attentive probe (brain.space + HBN) & \textbf{3.06} &
\textbf{5.11} & \textbf{0.85} \\
STST-JEPA + attentive probe (brain.space only) & 4.82 & 7.06 & 0.654 \\
Predict training mean (17.7 yr) & ≈ 10.09 & 13.27 & ≈ 0 \\
\end{longtable}

Relative to the predict-the-mean baseline, the best model reduces MAE by
69.7\% (3.06 vs 10.09 yr) and RMSE by 61.5\% (5.11 vs 13.27 yr). The
predict-the-mean MAE is approximated from the training age standard
deviation (\(\sigma = 12.65\)) via the Gaussian case identity
\(\mathbb{E}[|X - \mu|] = \sigma\sqrt{2/\pi}\); the MAE optimal
\emph{predict-the-median} baseline is tighter still (median ≈ 14 yr on
the training distribution → MAE ≈ 9.8 yr, also \(\approx 70\)\%
reduction). The RMSE baseline is the logged
\(\mathrm{rmse}_{\mathrm{true}}\) (13.268), which equals \(\sigma\) on a
mean centered label.

Figure 1 visualizes the headline result: predicted vs.~true age on the
3,367 held out validation sessions at the best evaluation step. The fit
(red) sits below the identity line (dashed) across most of the age
range, the characteristic regression-to-the-mean pattern of brain age
models that motivates the bias correction we apply later for downstream
BAG analyses.

\begin{figure}
\centering
\includegraphics[width=0.85\textwidth,height=\textheight]{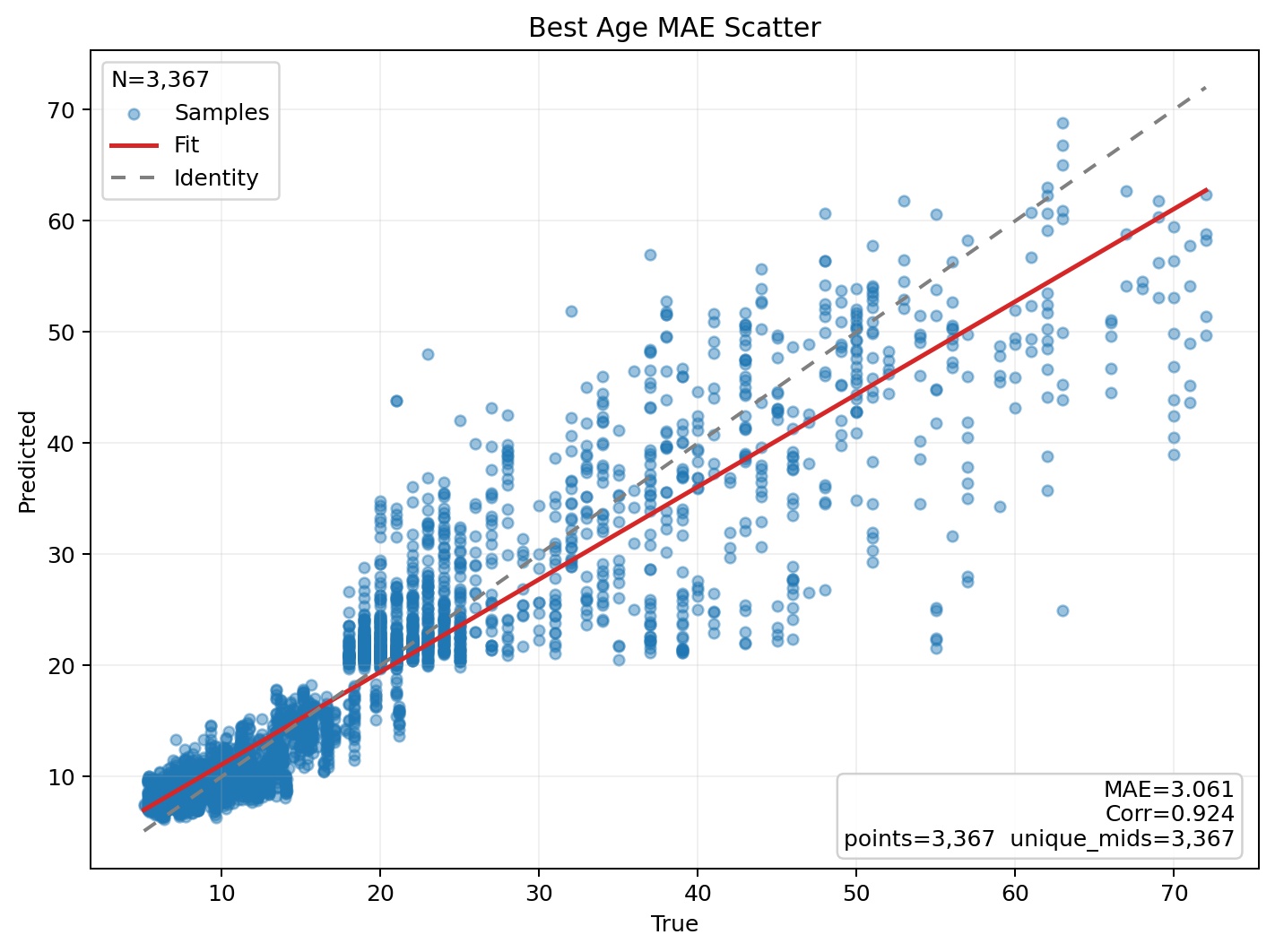}
\caption{\textbf{Figure 1 - Best predicted vs.~true age on the
held out validation set} (N = 3,367 sessions; MAE = 3.06 yr, Pearson r =
0.924). Identity line dashed; OLS fit in red.}
\end{figure}

\begin{figure}
\centering
\includegraphics[width=0.85\textwidth,height=\textheight]{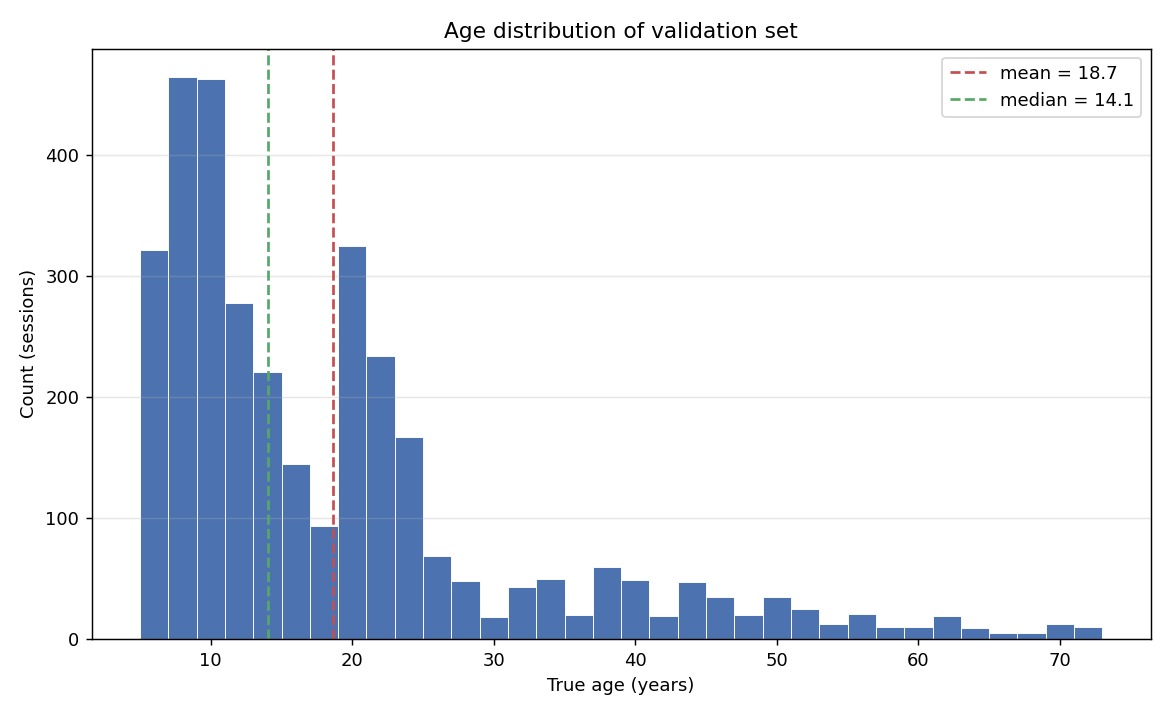}
\caption{\textbf{Figure 1b - True age distribution of the held out
validation evaluation set} (N = 3,367 sessions; mean 18.7 yr, median
14.1 yr). The distribution is pediatric heavy and right tailed - HBN
concentrates the mass in childhood and adolescence while brain.space
extends the tail through adulthood - which, together with the
compressed conditional age variance of such a cohort, is part of why the
combined absolute MAE is not directly comparable to adult clinical
benchmarks.}
\end{figure}

\hypertarget{pretraining-dynamics}{%
\subsection{Pretraining Dynamics}\label{pretraining-dynamics}}

Figure 2 shows the downstream validation age MAE trajectory across
pretraining on a shared step axis; both training loss terms (discussed
below) decline in concert.

\begin{figure}
\centering
\includegraphics[width=0.85\textwidth,height=\textheight]{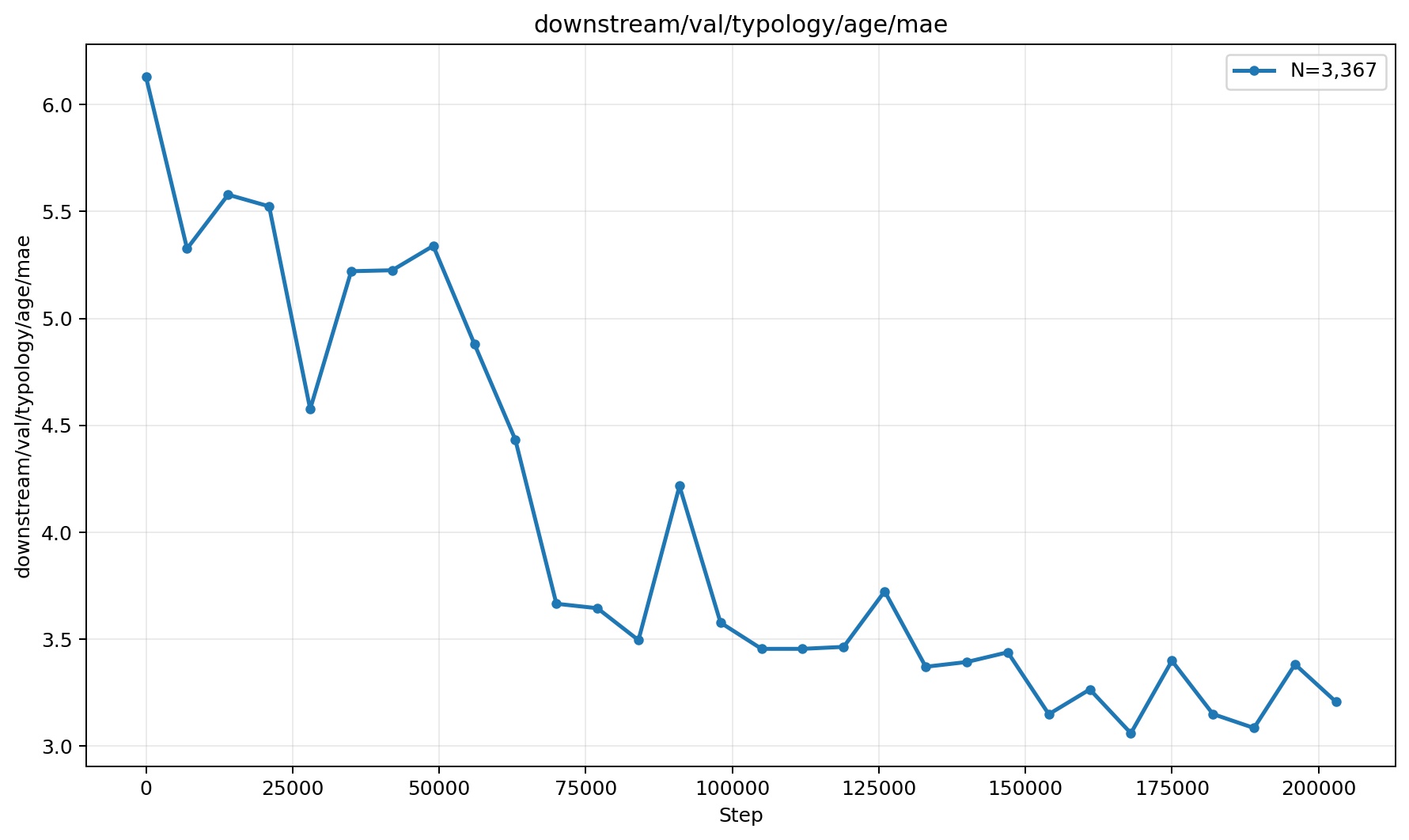}
\caption{\textbf{Figure 2 - Downstream validation age MAE as a
function of pretraining step} (N = 3,367 sessions per evaluation). MAE
drops from approximately 6.1 yr at step 0 to a best value near 3.06 yr
late in training.}
\end{figure}

\textbf{Loss decomposition.} The latent prediction term drops roughly
four orders of magnitude across training, from 0.0575 at step 49 to
\(1.93 \times 10^{-4}\) at step 155,549 (a 99.7\% reduction in linear
units, \textasciitilde2.5 decades in log units). The raw per patch
reconstruction term decreases more modestly - from 0.317 to 0.169 over
the same span (−47\%) - with a clear bend around steps 55,000--70,000
where it drops out of an early plateau. Late in training both terms
settle onto a noisy steady state plateau with the latent prediction loss
near \([1.5 \times 10^{-2}, 2.2 \times 10^{-2}]\) and the reconstruction
loss near \([0.13, 0.18]\).

The asymmetry between the two loss components is informative: the
latent prediction term is driven down as a primary objective, while the
reconstruction term behaves as a soft floor that stabilizes the latent
space rather than a co-minimized target. This is consistent with the
role we assigned to it in Methods - auxiliary regularization rather
than co-equal supervision. The model is trained to convergence (LR
schedule, EMA momentum warmup, and mask ratio statistics match the
configured values; weight-watcher spectral exponents are flat in late
training).

\hypertarget{auxiliary-downstream-probes}{%
\subsection{Auxiliary Downstream
Probes}\label{auxiliary-downstream-probes}}

Beyond age, the same attentive probe protocol was applied to two further
label families: a binary sex classifier and a panel of paradigm
classifiers. Table 3 summarizes each probe at its last evaluation,
together with a chance level baseline; balanced accuracy is reported
throughout (insensitive to class imbalance).

\textbf{Table 3 - Auxiliary probe performance on the held out
validation set.} ``Best'' values in parentheses for probes whose final
evaluation was not the trajectory maximum.

\begin{longtable}[]{@{}
  >{\raggedright\arraybackslash}p{(\columnwidth - 8\tabcolsep) * \real{0.3580}}
  >{\raggedright\arraybackslash}p{(\columnwidth - 8\tabcolsep) * \real{0.1111}}
  >{\raggedright\arraybackslash}p{(\columnwidth - 8\tabcolsep) * \real{0.1358}}
  >{\raggedleft\arraybackslash}p{(\columnwidth - 8\tabcolsep) * \real{0.2716}}
  >{\raggedleft\arraybackslash}p{(\columnwidth - 8\tabcolsep) * \real{0.1235}}@{}}
\toprule\noalign{}
\begin{minipage}[b]{\linewidth}\raggedright
Probe
\end{minipage} & \begin{minipage}[b]{\linewidth}\raggedright
Task
\end{minipage} & \begin{minipage}[b]{\linewidth}\raggedright
Metric
\end{minipage} & \begin{minipage}[b]{\linewidth}\raggedleft
Result
\end{minipage} & \begin{minipage}[b]{\linewidth}\raggedleft
Baseline
\end{minipage} \\
\midrule\noalign{}
\endhead
\bottomrule\noalign{}
\endlastfoot
\texttt{phenotype/sex} & binary & bal. acc. & 0.891 & 0.500 \\
\texttt{paradigm/eo\_ec} & binary & bal. acc. & 0.863 \emph{(best
0.882)} & 0.500 \\
\texttt{paradigm/nback} & 3 class & bal. acc. & 0.545 & 0.333 \\
\texttt{paradigm/mot} & 6 class & bal. acc. & 0.407 \emph{(best 0.432)}
& 0.167 \\
\texttt{paradigm/mmst} & 6 class & bal. acc. & 0.447 \emph{(best 0.482)}
& 0.167 \\
\end{longtable}

Numbers of classes are inferred from the first evaluation
balanced accuracy value (exactly \(1/K\) at initialization) and match
the label definitions in the dataset manifest.

\textbf{Above chance.} Sex reaches 0.891 balanced accuracy on the
validation set, climbing from 0.52 at the first evaluation. All four
paradigm classifiers exceed chance: \texttt{eo\_ec} (binary eyes open /
eyes closed) peaks at 0.88 balanced accuracy, \texttt{nback} (3 class)
reaches 1.6× chance, and the two 6 class paradigm probes (\texttt{mot},
\texttt{mmst}) reach 2.4×--2.9× chance. These results indicate that the
embeddings carry a representation of the subject's \emph{state} - what
paradigm or task phase a window was recorded under - and of at least
one coarse subject trait (sex). We do not characterize the multiclass
paradigm results as ``strong'' - \texttt{nback} at 0.545 is well above
chance (0.333) but still well below ceiling and should be read as
moderate.

\hypertarget{external-benchmark-performance}{%
\subsection{External Benchmark
Performance}\label{external-benchmark-performance}}

To complement the internal probe panel above, we evaluated our model on
the NeuralBench × brain.space EEG leaderboard, a public benchmark that
aggregates published models on a fixed held out test protocol. Three
tasks - binary sex, age regression, and a psychopathology composite
- admit direct comparison against the prior leaderboard entries.
Per task probes were trained from the same pretrained model under the
leaderboard's protocol (encoder lower layers held fixed, final layers
fine tuned on the leaderboard's training partition); the leaderboard
metric definitions are \texttt{test/bal\_acc} for sex and
\texttt{test/pearsonr} for age and psychopathology.

\textbf{Table 4 - NeuralBench × brain.space EEG leaderboard results
from a single shared model.} Our entries use 30 second input windows;
the standard NeuralBench protocol uses 2 second windows, so these
comparisons are not matched on input length (see the window length
caveat below).

\begin{longtable}[]{@{}
  >{\raggedright\arraybackslash}p{(\columnwidth - 10\tabcolsep) * \real{0.1429}}
  >{\raggedright\arraybackslash}p{(\columnwidth - 10\tabcolsep) * \real{0.1429}}
  >{\raggedleft\arraybackslash}p{(\columnwidth - 10\tabcolsep) * \real{0.1905}}
  >{\raggedright\arraybackslash}p{(\columnwidth - 10\tabcolsep) * \real{0.1429}}
  >{\raggedleft\arraybackslash}p{(\columnwidth - 10\tabcolsep) * \real{0.1905}}
  >{\raggedleft\arraybackslash}p{(\columnwidth - 10\tabcolsep) * \real{0.1905}}@{}}
\toprule\noalign{}
\begin{minipage}[b]{\linewidth}\raggedright
Task
\end{minipage} & \begin{minipage}[b]{\linewidth}\raggedright
Metric
\end{minipage} & \begin{minipage}[b]{\linewidth}\raggedleft
Ours (30 s)
\end{minipage} & \begin{minipage}[b]{\linewidth}\raggedright
Prior best (model)
\end{minipage} & \begin{minipage}[b]{\linewidth}\raggedleft
Margin
\end{minipage} & \begin{minipage}[b]{\linewidth}\raggedleft
Rank
\end{minipage} \\
\midrule\noalign{}
\endhead
\bottomrule\noalign{}
\endlastfoot
Sex & bal. acc. & \textbf{0.911} & 0.910 (ShallowFBCSPNet
{[}Schirrmeister et al., 2017{]}) & +0.001 & 1 / 18 \\
Age & Pearson r & \textbf{0.749} & 0.721 (REVE {[}Ouahidi et al.,
2025{]}) & +0.028 & 1 / 17 \\
Psychopath. & Pearson r & \textbf{0.215} & 0.137 (CBraMod {[}Wang et
al., 2025{]}) & +0.078 & 1 / 15 \\
\end{longtable}

\textbf{Window length caveat.} The Table 4 entries are produced with the
model's native 30 second input windows, whereas the standard NeuralBench
protocol evaluates on 2 second windows; the comparisons above are
therefore not matched on input length - the longer window gives our
model more context per prediction than the prior entries received -
and the margins should be read with that advantage in mind. Under the
standard 2 second protocol the same checkpoint remains rank 1 on sex
(balanced accuracy 0.913, tied leaderboard best) and on the
psychopathology composite (Pearson r 0.141, just above the prior best of
0.137, CBraMod), while its age correlation drops to Pearson r 0.691
(rank 4 of 17; the leaderboard's REVE entry leads at 0.721). The
30 second setting is thus what lifts age to rank 1, whereas sex and
psychopathology lead under either window length.

The sex result is best read as a tie with the prior leaderboard top
entry - a 0.001 absolute difference is well within run to run
variance.

For age regression, the +0.028 Pearson r margin over REVE corresponds to
a meaningful improvement on a metric on which entries on this
leaderboard have converged tightly. Reported numbers across the internal
split (MAE / RMSE) and the leaderboard (Pearson r) capture different
statistics on different held out partitions and should be read as
consistent indications of the same model from different angles.

For psychopathology, the absolute Pearson r of 0.215 is small - only
about 4.6\% of variance explained - but it is roughly 1.6× the prior
leaderboard best (0.137, CBraMod) under the same protocol. The
substantive takeaway is that the same model carries leaderboard leading
signal on a long horizon trait label under a public protocol, even
though the absolute correlation remains modest.

\hypertarget{brain-age-gap-and-behavioural-performance}{%
\subsection{Brain Age Gap and Behavioural
Performance}\label{brain-age-gap-and-behavioural-performance}}

A natural follow up question to the headline age prediction result is
whether the per session residual - the \textbf{brain age gap} (BAG =
predicted age − true age, after bias correction) - carries information
about behavioural performance beyond what is explained by chronological
age alone. Following the standard brain age convention, we
bias corrected the raw prediction by regressing it on true age in a
5 fold cross-fit over the union of validation and test partitions (N =
8,600 sessions, 2,109 subjects), giving a slope
\(\alpha = 0.642 \pm 0.006\) and intercept \(\beta = 8.65 \pm 0.15\).
The corrected BAG is age orthogonal by construction
(\(\mathrm{corr}(\mathrm{BAG}, \mathrm{age}) \approx 4 \times 10^{-5}\)),
which is why we lead with the \textbf{raw Pearson r} below: partialling
age a second time is a no-op, and the only role left for a partial
coefficient is as a defensive check against signal quality nuisance.

We correlated BAG against the 21 target behavioural capacity panel
defined in Methods §\emph{Behavioural Capacity Targets and Covariates}:
14 subject level capacity composites (broadcast across the subject's
recordings) and 7 per recording per task efficiency scores. The
partial r covariate set is a signal quality nuisance bundle only
(time-of-day, head circumference, nasin/temtem distances, hair
type/wash, glasses use, handedness); per the bundle's methodology, age /
sex / sleep / education are treated as \textbf{substrate, not
confounds}, and are not partialled out here.

\textbf{Table 5 - Top behavioural capacity correlates of the brain age
gap (linear bias correction; ordered by \textbar raw r\textbar).} Grain
``subject'' = per subject capacity composite broadcast to every
recording of that subject; ``recording'' = per task efficiency score,
one value per recording.

\begin{longtable}[]{@{}
  >{\raggedright\arraybackslash}p{(\columnwidth - 10\tabcolsep) * \real{0.4824}}
  >{\raggedright\arraybackslash}p{(\columnwidth - 10\tabcolsep) * \real{0.1294}}
  >{\raggedleft\arraybackslash}p{(\columnwidth - 10\tabcolsep) * \real{0.0824}}
  >{\raggedleft\arraybackslash}p{(\columnwidth - 10\tabcolsep) * \real{0.0941}}
  >{\raggedleft\arraybackslash}p{(\columnwidth - 10\tabcolsep) * \real{0.1294}}
  >{\raggedleft\arraybackslash}p{(\columnwidth - 10\tabcolsep) * \real{0.0824}}@{}}
\toprule\noalign{}
\begin{minipage}[b]{\linewidth}\raggedright
Target
\end{minipage} & \begin{minipage}[b]{\linewidth}\raggedright
Grain
\end{minipage} & \begin{minipage}[b]{\linewidth}\raggedleft
n
\end{minipage} & \begin{minipage}[b]{\linewidth}\raggedleft
r
\end{minipage} & \begin{minipage}[b]{\linewidth}\raggedleft
partial r
\end{minipage} & \begin{minipage}[b]{\linewidth}\raggedleft
z
\end{minipage} \\
\midrule\noalign{}
\endhead
\bottomrule\noalign{}
\endlastfoot
Stroop efficiency & recording & 1,492 & −0.089 & −0.062 & −3.45 \\
Stroop efficiency (composite) & subject & 7,366 & −0.080 & −0.069 &
−6.89 \\
Dual oddball efficiency & recording & 987 & −0.071 & −0.098 & −2.23 \\
n-back efficiency (composite) & subject & 7,993 & −0.057 & −0.048 &
−5.11 \\
General efficiency & subject & 8,436 & −0.049 & −0.044 & −4.54 \\
MOT efficiency & recording & 1,189 & −0.049 & −0.016 & −1.68 \\
n-back efficiency & recording & 1,729 & −0.046 & −0.043 & −1.91 \\
General efficiency (age residualised) & subject & 8,436 & −0.045 &
−0.048 & −4.17 \\
Attention control & subject & 8,468 & −0.043 & −0.045 & −3.96 \\
Flanker efficiency (composite) & subject & 4,414 & −0.034 & −0.036 &
−2.24 \\
P3 efficiency (composite) & subject & 7,447 & −0.033 & −0.024 & −2.82 \\
\end{longtable}

\textbf{Pattern of negative correlation between the brain age gap and
cognitive performance.} Every target with above chance information shows
a \textbf{negative} correlation: a larger (older looking) brain age gap
tracks worse speeded cognitive performance. The single largest
coefficient is per recording Stroop efficiency at r = −0.089 on its
native 1,492 recording sample (z = −3.45), and the highest powered
association is the subject level Stroop composite at r = −0.080 over n =
7,366 (z = −6.89). The most informative ``EEG beyond age'' association
is the age residualised general efficiency target (r = −0.045, n =
8,436): this target is itself age residualised by construction, so any
nonzero correlation with the age orthogonal BAG is behavioural signal
that survives a double accounting for age. The effects are small (all
\textbar r\textbar{} \textless{} 0.10), as is typical of brain age gap ×
behaviour associations in healthy cohorts of this scale; the
load bearing observation is the \textbf{direction-of-effect consistency}
across the panel, not any individual coefficient.

\textbf{n-inflation caveat.} Subject level composites repeat a single
per subject value across all of that subject's recordings, which
inflates the effective sample size of the correlation relative to the
number of statistically independent observations (2,109 subjects, not
8,600 recordings). The per recording rows in Table 5 give honest n, but
at a substantially smaller cohort (a few hundred to \textasciitilde1,700
recordings per task). The subject level z-statistics should be read as
upper bounds on the true significance, and the per recording rows as the
conservative floor. The two grains are directionally concordant in every
task family they both cover (Stroop, n-back, MOT, P3, Flanker,
dual oddball, MMST).

\textbf{Sensitivity to bias correction order.} Re-running the same
analysis with a quadratic bias correction in place of the linear one
(i.e., regressing prediction on a 2nd order polynomial in true age)
leaves the answer numerically unchanged: the largest
\textbar Δr\textbar{} across all 21 targets is 0.010. On this cohort the
quadratic coefficient is near zero, so we report the linear correction
in the body and treat the quadratic as a sensitivity check.

\textbf{Partial r vs raw r movement.} Across the 11 reported targets,
the partial r column (controlling for the signal quality nuisance
bundle) stays within 0.03 of the raw r for nine of them, with the two
largest movements at per recording dual oddball efficiency (−0.071 →
−0.098) and per recording MOT efficiency (−0.049 → −0.016). The nuisance
bundle neither inflates nor systematically erodes the headline pattern;
the negative direction story survives at both grains.

\textbf{Caveat: task window contribution to BAG.} The pretraining input
mixes resting (\texttt{eo\_ec}) and task (\texttt{mot}) paradigms
(Methods §\emph{EEG Segmentation and Preprocessing}), so a fraction of
each session's age prediction is computed from task evoked dynamics. The
BAG residual therefore carries some task state information by
construction, and its negative correlation with task efficiency targets
above could in principle be amplified through this within window
channel. A resting state only BAG replication is the cleanest control
and is left to follow up work.

Together these correlations constitute a small but consistent
behavioural validation of the brain age gap on this corpus: BAG is not
just an age prediction residual with no external referent - it tracks
speeded cognitive performance in the expected direction across a
heterogeneous task panel, while remaining far below the magnitude
required to interpret as a clinical biomarker. Figure 3 plots all 21
targets as a forest of raw r with 95\% confidence intervals.

\begin{figure}
\centering
\includegraphics[width=0.85\textwidth,height=\textheight]{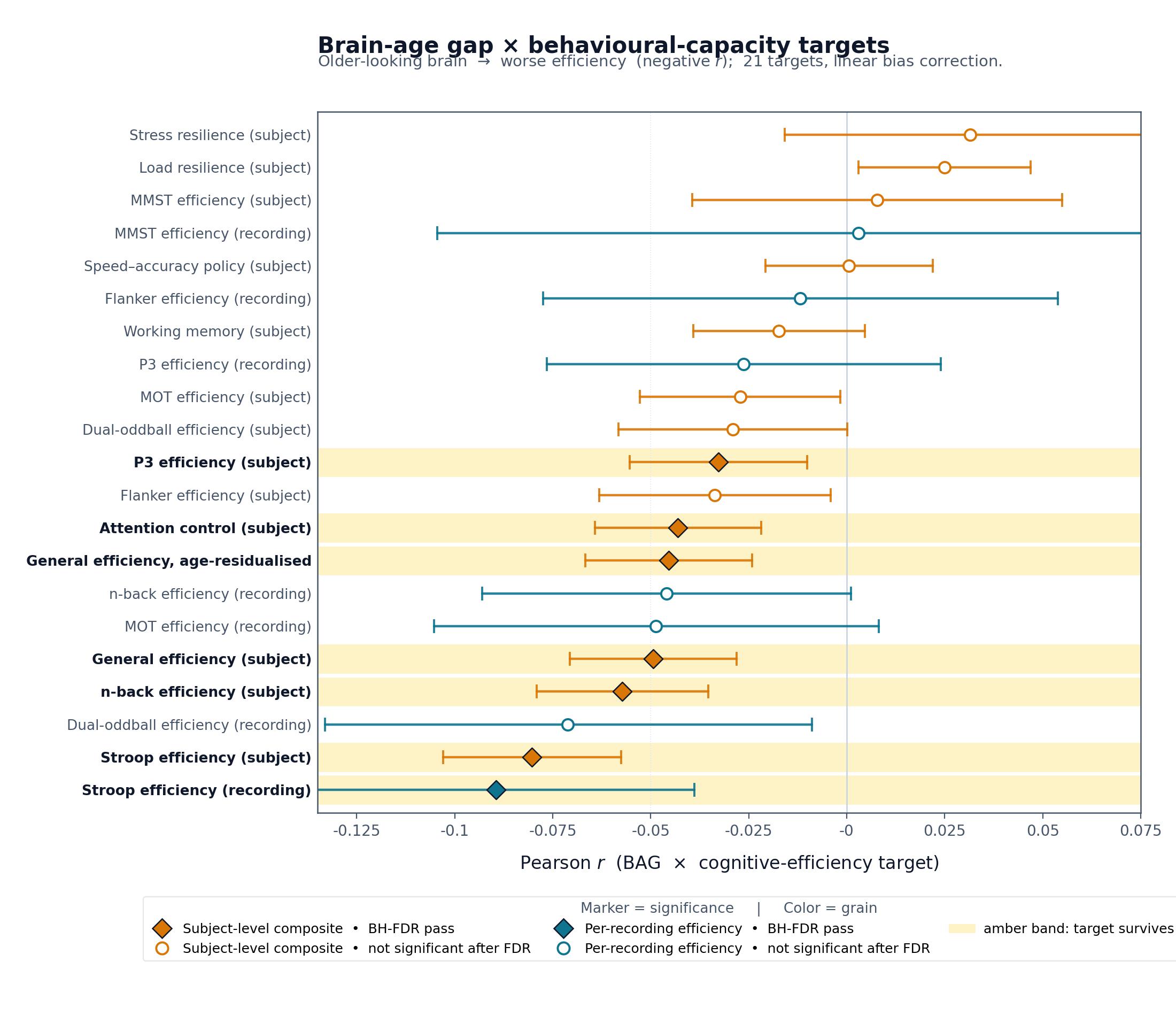}
\caption{\textbf{Figure 3 - Brain age gap × behavioural capacity
targets (forest plot, linear bias correction).} Each row is one target;
markers are Pearson r and whiskers are Fisher 95\% CIs. Filled diamonds
= BH-FDR-pass at q = 0.05 (amber banded rows); hollow circles = not
significant after FDR. Subject level capacity composites (orange) and
per recording task efficiency scores (teal) are shown together. Every
FDR significant target points in the negative direction (larger
brain age gap → worse cognitive efficiency); 7 of the 21 targets survive
BH-FDR at q = 0.05.}
\end{figure}

\hypertarget{discussion}{%
\section{Discussion}\label{discussion}}

The 3.06 year held out validation MAE reported above is the floor of the
\emph{probe protocol} we ran, not the floor of the \emph{representation}
we learned. Three orthogonal levers remain unexercised in this paper:
scaling the pretraining corpus, modifying the supervision the encoder
receives during pretraining, and changing the downstream adaptation
protocol from frozen probe to end-to-end fine tuning. A serious next
iteration of STST-JEPA should exercise all three - they target
different parts of the bias--variance budget and are at minimum
additive, plausibly complementary.

\textbf{Lever 1 - pretraining corpus scaling.} The internal roadmap
targets roughly 3× the current corpus (\textasciitilde143,000 sessions,
up from 47,703). We expect sublinear gains: latent prediction SSL in
adjacent domains {[}I-JEPA, Assran et al., 2023; V-JEPA, Bardes et al.,
2024; data2vec, Baevski et al., 2022{]} and the in EEG corpus size
curves implied by LaBraM {[}Jiang et al., 2024{]} show
representation level gains accruing on a log axis. We do not claim a
\emph{biological} age MAE floor - chronological age is decodable from
EEG content in principle and the only true lower bound is 0 - but a
\emph{representational} and \emph{session} floor remains: how much
age discriminative signal a 30 second EEG window can carry, recovered
through a tractable encoder and decoded by a finite capacity probe. As a
falsifiable hypothesis, we project a 3×-corpus MAE in the \textbf{2.6 --
2.9 yr range} with proportionally smaller RMSE - a modest improvement
on the headline number, calibrated by the log axis scaling priors above
rather than by an asserted biology floor. The marginal data is more
likely to move dials we cannot yet move on this corpus: per source MAE /
RMSE breakdowns, tighter NeuralBench margins where headroom remains
substantial, and stability of probes whose current trajectories are
noisy at this scale. We conjecture that the NeuralBench psychopathology
endpoint (Pearson r = 0.215, R² ≈ 4.6\%) is the one most likely to
benefit from corpus scaling - it has by far the most headroom of the
three leaderboard tasks, whereas sex is already at the ShallowFBCSPNet
tie and age has a tight margin on Pearson r.

\textbf{Lever 2 - auxiliary losses at pretraining time.} The
pretraining objective never sees age, so any age discriminative geometry
in the embedding is incidental to the latent prediction + reconstruction
objective. Three additions are cheap and orthogonal: (i) a light
age regression auxiliary head on the pooled inducing query token at
\(\lambda_{\mathrm{age}} \approx 0.05\) (anchor grid
\(\{0.01, 0.05, 0.2\}\)), keeping the SSL terms dominant while biasing
the representation toward age relevant axes; (ii) a learned cohort /
source / montage token prefixed to the encoder so the SSL target no
longer wastes capacity reinventing brain.space-vs-HBN separation already
discernible from coordinates; (iii) a cohort balanced
contrastive-by-age-band tertiary loss at small weight, shaping the
latent geometry to put far apart ages far apart and
same-band-different-subject pairs close. Each is an ablation, not a
guaranteed win; the joint objective ordering is empirically
unpredictable.

\textbf{Lever 3 - fine tuning options at downstream time.} All numbers
reported in §Validation Age Regression come from a \textbf{frozen}
backbone; the simplest next step is end-to-end fine tuning under age
supervision. Vision-SSL precedent suggests a linear probe → end-to-end
gap of 1 -- 3 percentage points; relative to the current frozen probe
regime - where some of the residual variance is plausibly attributable
to the probe's limited capacity rather than to the representation -
this maps to a plausible \textbf{0.2 -- 0.5 yr MAE reduction}, taken as
a falsifiable projection, not a deliverable. Where compute or stability
is the constraint, \textbf{layer wise LR decay} (e.g., factor 0.65 per
depth, freezing the lowest 12 of 24 layers) and \textbf{LoRA-style
adapters} (rank 8 -- 16 on attention QKV projections) are the standard
parameter efficient escape valves and would preserve the SSL
representation that already wins on the NeuralBench multitask panel.
Two further options exploit structure already in the data: a
\textbf{subject level mean teacher} distillation during fine tuning,
leveraging the multiple sessions per subject in the brain.space finetune
cohort (val + test union: 8,600 sessions across 2,109 subjects, ≈ 4.1
sessions/subject), and \textbf{multitask fine tuning} (age + sex +
NeuralBench psychopathology jointly), where the multitask EEG
literature suggests the weakest endpoint typically gains most - making
psychopathology the natural conjectured beneficiary.

None of the above is reported empirically in this paper. We list them to
be explicit about where the next gains are expected to come from: Lever
1 addresses the data side of the bias--variance curve, while Levers 2
and 3 address the supervision side at pretraining time and at
downstream adaptation time respectively. The present whitepaper reports
the deliberately conservative point in their joint design space -
frozen backbone, fixed SSL objective, no age supervision during
pretraining. All projected numbers above are falsifiable hypotheses to
be tested in follow up work, not promised deliverables, and they do not
change the abstract's standing caveats: no controlled joint objective
ablation, no fixed protocol evaluation on our internal brain.space test
partition (the NeuralBench results are fixed protocol test evaluations),
and no clinical biomarker claim.

\hypertarget{acknowledgements}{%
\section{Acknowledgements}\label{acknowledgements}}

This work was supported by the Israel Innovation Authority (IIA) grant
84608. We thank Tom Touati and the Brain.Space research group for their
insights and suggestions.

\hypertarget{references}{%
\section{References}\label{references}}

\begin{enumerate}
\def\labelenumi{\arabic{enumi}.}
\tightlist
\item
  Y. LeCun and Courant, ``A Path Towards Autonomous Machine
  Intelligence,'' 2022. Accessed: Jul.~06, 2026. {[}Online{]}.
  Available:
  https://www.semanticscholar.org/paper/A-Path-Towards-Autonomous-Machine-Intelligence-LeCun-Courant/775f42ed458b8c5b0f2094ea4ff5b64c557b1a34
\item
  D. A. Engemann et al., ``A reusable benchmark of brain-age prediction
  from M/EEG resting-state signals,'' NeuroImage, vol.~262, p.~119521,
  Nov.~2022, doi: 10.1016/j.neuroimage.2022.119521.
\item
  T. Chen, S. Kornblith, M. Norouzi, and G. Hinton, ``A Simple Framework
  for Contrastive Learning of Visual Representations,'' Jul.~01, 2020,
  arXiv: arXiv:2002.05709. doi: 10.48550/arXiv.2002.05709.
\item
  L. M. Alexander et al., ``An open resource for transdiagnostic
  research in pediatric mental health and learning disorders,'' Sci
  Data, vol.~4, no. 1, p.~170181, Dec.~2017, doi:
  10.1038/sdata.2017.181.
\item
  A. Vaswani et al., ``Attention Is All You Need,'' Aug.~02, 2023,
  arXiv: arXiv:1706.03762. doi: 10.48550/arXiv.1706.03762.
\item
  D. P. Kingma and M. Welling, ``Auto-Encoding Variational Bayes,''
  Dec.~10, 2022, arXiv: arXiv:1312.6114. doi: 10.48550/arXiv.1312.6114.
\item
  D. Kostas, S. Aroca-Ouellette, and F. Rudzicz, ``BENDR: using
  transformers and a contrastive self-supervised learning task to learn
  from massive amounts of EEG data,'' Jan.~28, 2021, arXiv:
  arXiv:2101.12037. doi: 10.48550/arXiv.2101.12037.
\item
  C. Yang, M. Westover, and J. Sun, ``BIOT: Biosignal Transformer for
  Cross-data Learning in the Wild,'' in Advances in Neural Information
  Processing Systems, Curran Associates, Inc., 2023, pp.~78240--78260.
  Accessed: Jul.~06, 2026. {[}Online{]}. Available:
  https://proceedings.neurips.cc/paper\_files/paper/2023/hash/f6b30f3e2dd9cb53bbf2024402d02295-Abstract-Conference.html
\item
  J.-B. Grill et al., ``Bootstrap your own latent: A new approach to
  self-supervised Learning,'' Sep.~10, 2020, arXiv: arXiv:2006.07733.
  doi: 10.48550/arXiv.2006.07733.
\item
  H. Sun et al., ``Brain age from the electroencephalogram of sleep,''
  Neurobiol Aging, vol.~74, pp.~112--120, Feb.~2019, doi:
  10.1016/j.neurobiolaging.2018.10.016.
\item
  C. Wang et al., ``BrainBERT: Self-supervised representation learning
  for intracranial recordings,'' Feb.~28, 2023, arXiv: arXiv:2302.14367.
  doi: 10.48550/arXiv.2302.14367.
\item
  J. Wang et al., ``CBraMod: A Criss-Cross Brain Foundation Model for
  EEG Decoding,'' Nov.~06, 2025, arXiv: arXiv:2412.07236. doi:
  10.48550/arXiv.2412.07236.
\item
  A. Baevski, W.-N. Hsu, Q. Xu, A. Babu, J. Gu, and M. Auli, ``data2vec:
  A General Framework for Self-supervised Learning in Speech, Vision and
  Language,'' Oct.~25, 2022, arXiv: arXiv:2202.03555. doi:
  10.48550/arXiv.2202.03555.
\item
  R. T. Schirrmeister et al., ``Deep learning with convolutional neural
  networks for EEG decoding and visualization,'' Hum Brain Mapp,
  vol.~38, no. 11, pp.~5391--5420, Nov.~2017, doi: 10.1002/hbm.23730.
\item
  N. M. Foumani, G. Mackellar, S. Ghane, S. Irtza, N. Nguyen, and M.
  Salehi, ``EEG2Rep: Enhancing Self-supervised EEG Representation
  Through Informative Masked Inputs,'' Jun.~18, 2024, arXiv:
  arXiv:2402.17772. doi: 10.48550/arXiv.2402.17772.
\item
  V. J. Lawhern, A. J. Solon, N. R. Waytowich, S. M. Gordon, C. P. Hung,
  and B. J. Lance, ``EEGNet: a compact convolutional neural network for
  EEG-based brain-computer interfaces,'' J Neural Eng, vol.~15, no. 5,
  p.~056013, Oct.~2018, doi: 10.1088/1741-2552/aace8c.
\item
  J. Zhou et al., ``iBOT: Image BERT Pre-Training with Online
  Tokenizer,'' Jan.~27, 2022, arXiv: arXiv:2111.07832. doi:
  10.48550/arXiv.2111.07832.
\item
  W.-B. Jiang, L.-M. Zhao, and B.-L. Lu, ``Large Brain Model for
  Learning Generic Representations with Tremendous EEG Data in BCI,''
  May 29, 2024, arXiv: arXiv:2405.18765. doi: 10.48550/arXiv.2405.18765.
\item
  R. Balestriero and Y. LeCun, ``LeJEPA: Provable and Scalable
  Self-Supervised Learning Without the Heuristics,'' Nov.~14, 2025,
  arXiv: arXiv:2511.08544. doi: 10.48550/arXiv.2511.08544.
\item
  K. He, X. Chen, S. Xie, Y. Li, P. Dollár, and R. Girshick, ``Masked
  Autoencoders Are Scalable Vision Learners,'' Dec.~19, 2021, arXiv:
  arXiv:2111.06377. doi: 10.48550/arXiv.2111.06377.
\item
  W. Cui et al., ``Neuro-GPT: Towards A Foundation Model for EEG,''
  Mar.~02, 2024, arXiv: arXiv:2311.03764. doi:
  10.48550/arXiv.2311.03764.
\item
  R. Xiong et al., ``On Layer Normalization in the Transformer
  Architecture,'' Jun.~29, 2020, arXiv: arXiv:2002.04745. doi:
  10.48550/arXiv.2002.04745.
\item
  J. H. Cole and K. Franke, ``Predicting Age Using Neuroimaging:
  Innovative Brain Ageing Biomarkers,'' Trends Neurosci, vol.~40, no.
  12, pp.~681--690, Dec.~2017, doi: 10.1016/j.tins.2017.10.001.
\item
  R. P. N. Rao and D. H. Ballard, ``Predictive coding in the visual
  cortex: a functional interpretation of some extra-classical
  receptive-field effects,'' Nat Neurosci, vol.~2, no. 1, pp.~79--87,
  Jan.~1999, doi: 10.1038/4580.
\item
  Y. E. Ouahidi et al., ``REVE: A Foundation Model for EEG -- Adapting
  to Any Setup with Large-Scale Pretraining on 25,000 Subjects,''
  Oct.~24, 2025, arXiv: arXiv:2510.21585. doi:
  10.48550/arXiv.2510.21585.
\item
  A. Bardes et al., ``Revisiting Feature Prediction for Learning Visual
  Representations from Video,'' Feb.~15, 2024, arXiv: arXiv:2404.08471.
  doi: 10.48550/arXiv.2404.08471.
\item
  J. Su, Y. Lu, S. Pan, A. Murtadha, B. Wen, and Y. Liu, ``RoFormer:
  Enhanced Transformer with Rotary Position Embedding,'' Nov.~08, 2023,
  arXiv: arXiv:2104.09864. doi: 10.48550/arXiv.2104.09864.
\item
  M. Assran et al., ``Self-Supervised Learning from Images with a
  Joint-Embedding Predictive Architecture,'' Apr.~13, 2023, arXiv:
  arXiv:2301.08243. doi: 10.48550/arXiv.2301.08243.
\item
  J. Lee, Y. Lee, J. Kim, A. R. Kosiorek, S. Choi, and Y. W. Teh, ``Set
  Transformer: A Framework for Attention-based Permutation-Invariant
  Neural Networks,'' May 26, 2019, arXiv: arXiv:1810.00825. doi:
  10.48550/arXiv.1810.00825.
\item
  K. Franke and C. Gaser, ``Ten Years of BrainAGE as a Neuroimaging
  Biomarker of Brain Aging: What Insights Have We Gained?,'' Front
  Neurol, vol.~10, p.~789, 2019, doi: 10.3389/fneur.2019.00789.
\item
  K. Friston, ``The free-energy principle: a unified brain theory?,''
  Nat Rev Neurosci, vol.~11, no. 2, pp.~127--138, Feb.~2010, doi:
  10.1038/nrn2787.
\item
  Z. W. Pylyshyn and R. W. Storm, ``Tracking multiple independent
  targets: evidence for a parallel tracking mechanism,'' Spat Vis,
  vol.~3, no. 3, pp.~179--197, 1988, doi: 10.1163/156856888x00122.
\item
  A. Bardes, J. Ponce, and Y. LeCun, ``VICReg:
  Variance-Invariance-Covariance Regularization for Self-Supervised
  Learning,'' Jan.~28, 2022, arXiv: arXiv:2105.04906. doi:
  10.48550/arXiv.2105.04906.
\item
  A. Clark, ``Whatever next? Predictive brains, situated agents, and the
  future of cognitive science,'' Behav Brain Sci, vol.~36, no. 3,
  pp.~181--204, Jun.~2013, doi: 10.1017/S0140525X12000477.
\end{enumerate}

\end{document}